\documentclass{article} 
\usepackage{iclr2027_conference,times}

\usepackage{amsmath,amsfonts,bm}

\def\1{\bm{1}}

\DeclareMathAlphabet{\mathsfit}{\encodingdefault}{\sfdefault}{m}{sl}
\SetMathAlphabet{\mathsfit}{bold}{\encodingdefault}{\sfdefault}{bx}{n}

\usepackage{hyperref}
\hypersetup{hidelinks}
\usepackage{url}
\usepackage{booktabs}
\usepackage{multirow}
\usepackage{graphicx}
\usepackage{wrapfig}
\usepackage{amssymb}
\usepackage{algorithm}
\usepackage{algpseudocode}

\usepackage{amsthm}

\theoremstyle{plain}

\newcommand{\modelname}{\texttt{BQ-LoRA}}
\newcommand{\moduleA}{BQB}
\newcommand{\moduleB}{DPC}

\title{Behavior Quotient Learning for Low-Rank Adaptation of LLM Agents}

\author{\textbf{Pengyang Zhou\thanks{Equal contribution.}\hspace{0.3em}, Xiaobin Tu\footnotemark[1]\hspace{0.3em}, Zhengxi Liu, Rongkun Xue, Haochen Li,}\\
\textbf{Miancan Liu, Ziyuan Chen, Yinggui Wang\thanks{Corresponding author: \href{mailto:wyinggui@gmail.com}{wyinggui@gmail.com}}\hspace{0.3em}, Jinkui Ren, Xiantao Zhang}\\
Alibaba Cloud}

\iclrpreprintcopy
\begin{document}

\maketitle
\vspace{-18pt}

\begin{abstract}
LLM-based agents rely on heterogeneous interaction capabilities to accomplish complex tasks.
Existing approaches often distribute these capabilities across multiple LoRA adapters, which increases adapter storage requirements and introduces routing overhead during inference.
A single LoRA avoids this overhead, but learning from diverse agent trajectories under a fixed rank budget presents two challenges.
First, trajectories with different interaction traces and parameter gradients can induce equivalent changes in decision distributions, causing repeated updates to overemphasize redundant behavioral changes.
Second, an aggregated update may exceed the rank budget of the adapter, and approximating it in weight space can distort the decision changes that it is intended to produce.
We propose \modelname, a low-rank adaptation framework that organizes trajectory updates through a local behavior quotient manifold.
It contains two modules, i.e., behavior quotient balancing (\moduleA) and decision preserving compression (\moduleB).
\moduleA~constructs the quotient manifold from decision distributions and reweights trajectory update directions according to their local density in the quotient tangent space.
\moduleB~projects the balanced gradient onto the intrinsic fixed rank tangent space and refactorizes the resulting target by jointly controlling effective weight error and distortion of decision distributions.
Experiments on AppWorld and BrowseComp-Plus compare \modelname~with standard LoRA and recent low-rank adaptation methods, while separate ablations evaluate the complementary contributions of both components.
\end{abstract}

\section{Introduction}

Agents built on large language models (LLMs) interact with external environments to accomplish specified goals~\citep{ReAct,AgentSurvey}.
Despite broad capabilities, LLM-based agents often require task-specific training to operate effectively in new interaction settings~\citep{AgentTuning,AgentScaler}.
Low-Rank Adaptation (LoRA)~\citep{LoRA} offers an efficient alternative to full-model fine-tuning by learning factorized updates to frozen backbone weights.

LoRA-based agent adaptation broadly follows two designs.
The first assigns different agent capabilities to \textbf{multiple adapters}.
In multi-agent workflows, separate LoRA groups support reasoning, execution, and summarization~\citep{MoRAgent}.
Within tool-use workflows, adapter specialization separates tool selection from argument generation~\citep{DualTune} and reasoning from tool use~\citep{DART}.
More recent work trains a dedicated adapter for each identified capability gap, with inference-time routing selecting the relevant adapter~\citep{TRACE}.
Managing multiple specialized adapters requires coordination during inference.
Adapter-dependent states can also increase serving costs in multi-agent workflows~\citep{LRAgent}.
The second avoids adapter-management overhead by learning a shared low-rank update with a \textbf{single adapter}.
For agent adaptation, the shared update is learned directly from interaction trajectories~\citep{FireAct,SMART,AndroidGen}.
This design incorporates supervision for reasoning and action generation into a common set of low-rank parameters, allowing the same adapter to support the agent across successive interaction steps.
More generally, single-adapter LoRA variants improve the parameterization and optimization of low-rank updates to make more effective use of a limited trainable parameter budget~\citep{LoRA-S,StelLA,RiemannianLoRA}.

However, behavioral redundancy among trajectory updates can limit how effectively a single fixed-rank adapter learns from diverse supervision.
Behavioral redundancy arises when different trajectory updates induce similar changes in decision representations at the same reference histories.
Trajectory synthesis broadens interaction coverage~\citep{TDScaling}, while trajectory curation improves supervision quality~\citep{CurateEvo}.
Training utility is also assessed through model-dependent signals~\citep{RODS,LESS,DataInf} and critical-step analysis~\citep{ATLaS,CSO}.
These methods identify useful supervision but do not compare the behavioral effects of different trajectory updates.
Behavioral equivalence can occur despite substantial differences in trajectory text and parameter gradients.
Under a fixed rank budget, repeated equivalent updates can disproportionately shape the learned adapter and leave updates with distinct effects on these reference decisions underrepresented.
This gap raises a central question.
\textit{\textbf{How can a single fixed-rank LoRA learn diverse agent behaviors?}}

Addressing this question presents two challenges.
Firstly, \textit{how to identify behaviorally equivalent trajectory updates?} (\textbf{CH1})
Recorded trajectories alone do not reveal whether their training updates are behaviorally equivalent.
Different interaction traces and parameter gradients may still induce similar changes in decision representations.
TopoCurate~\citep{TopoCurate} estimates semantic equivalence from similarities between recorded tool actions and environment observations.
However, this criterion does not compare how trajectory updates change decision representations.
Identifying behavioral equivalence therefore requires comparing these changes at the same reference histories.
Secondly, \textit{how to preserve changes in decision representations under a fixed LoRA rank constraint?} (\textbf{CH2})
Diverse trajectory updates can produce a desired effective update whose rank exceeds this constraint.
Mapping such an update to a single fixed-rank adapter necessarily introduces approximation error.
Weight-space error alone does not capture how compression distorts these changes.
A suitable fixed-rank representative must preserve the effects of the target on decision representations and remain invariant to equivalent LoRA factorizations.

To address these challenges, we propose \modelname, a framework that learns diverse agent behaviors through a single fixed-rank LoRA adapter.
Our key idea is to use a shared geometry of decision behavior to guide both trajectory weighting and low-rank compression.
For \textbf{CH1}, we develop behavior quotient balancing (\moduleA), which compares trajectory update directions through their first-order effects on normalized decision representations at fixed interaction histories.
This comparison identifies locally equivalent behavioral changes while disregarding parameter differences that are invisible to the reference decisions.
\moduleA~then reweights the directions according to their local concentration, reducing repeated contributions and giving greater relative influence to distinct behavioral changes.
For \textbf{CH2}, we develop decision preserving compression (\moduleB), which projects the balanced direction onto the tangent space of the current fixed-rank adapter.
The resulting target depends on the effective weights and remains invariant to equivalent LoRA factorizations.
\moduleB~compresses this target by jointly controlling weight approximation error and first-order decision distortion in the same behavior geometry.
Together, the two modules connect behavioral coverage during training with preservation under the rank constraint, yielding one adapter that can be deployed directly.

We summarize our contributions as follows:
(1) We formalize behavioral equivalence among updates through a local quotient geometry and introduce \moduleA~to balance trajectory contributions according to their effects on agent decisions.
(2) We develop \moduleB, which combines a factorization-invariant tangent projection with compression that accounts for both weight error and first-order decision distortion under the original LoRA rank budget.
(3) We evaluate \modelname~on AppWorld and BrowseComp-Plus with two backbone sizes, comparing performance and interaction counts against standard adaptation, single-adapter optimization, adapter specialization, and trajectory curation.

\section{Related Work}

\subsection{Low-Rank Adaptation}
Low-rank adaptation (LoRA) freezes pretrained weights and parameterizes task-specific updates using trainable low-rank factors~\citep{LoRA}.
Existing LoRA variants improve adaptation through informative initialization~\citep{PiSSA,LoRA-GA,LoRA-One,LoRAM,FILet,LoRA-DA},
adaptive rank allocation~\citep{AdaLoRA,GoRA,EVA,IGU-LoRA,FlexLoRA,RaLoRA},
modified training procedures~\citep{LoRA+,LoRA-Pro,muA},
and input-conditioned utilization of the low-rank subspace~\citep{FouRA,TopLoRA,U-LoRA,FlyLoRA}.
Geometry-aware methods further improve LoRA optimization through Riemannian or transformation-invariant preconditioning~\citep{RiemannianLoRA,LoRA-RITE}, factor refactorization and balancing~\citep{RefLoRA,BalancedLoRA}, and manifold-based formulations~\citep{LoRA-S,Riemannion,StelLA,PoLAR}.
Multi-LoRA approaches use multiple low-rank parameter sets.
General-purpose research studies the routing and composition of LoRA experts for multi-task adaptation and adapter reuse~\citep{LoRAHub,MoLE,LoRAMoE,AdaMoLE,HMoRA,MeteoRA,LoRA-Mixer,LD-MoLE}.
Recent work on agentic systems uses multiple specialized adapters to model heterogeneous capabilities and coordinate their activation across agent trajectories~\citep{MoRAgent,VideoMind,DualTune,DART,TRACE}.

\subsection{Trajectory Redundancy}
Agent trajectories with different interaction traces may provide overlapping training signals, making trajectory redundancy an important consideration in agent post-training.
Existing work has primarily addressed this issue as a data-quality problem, using diversity-aware synthesis~\citep{TDScaling} and failure-driven curation~\citep{CurateEvo} to improve the composition of training data before optimization.
Moving from data composition to training effects, prior work estimates sample utility from optimization signals~\citep{RODS,LESS,DataInf} and localizes useful supervision through critical-step selection~\citep{ATLaS,CSO}.
These criteria identify useful supervision at different levels, but they do not determine whether distinct trajectory-induced updates produce equivalent changes in agent behavior.
TopoCurate~\citep{TopoCurate} examines this equivalence through a semantic quotient topology based on embedding similarities between tool actions and environment observations.
Because this relation is defined over recorded interaction content, TopoCurate does not determine whether different training updates induce the same behavioral change.
Interface perturbation studies further show that successful task execution does not necessarily reflect semantic tool-use competence~\citep{TrajectorySFT}.
Our work defines trajectory redundancy through update-induced behavioral changes, using the resulting equivalence to balance supervision before representing the learned update with a single fixed-rank LoRA.

\section{Preliminaries}

A quotient manifold is obtained by identifying equivalent points in a smooth total space~\citep{QuotientManifold}.
Let $\overline{\mathcal M}$ be a smooth manifold equipped with an equivalence relation $\sim$.
For $x\in\overline{\mathcal M}$, its equivalence class and the resulting quotient set are defined as:
\begin{equation}
[x]=\left\{y\in\overline{\mathcal M}:y\sim x\right\},\qquad \mathcal M=\overline{\mathcal M}/\sim=\left\{[x]:x\in\overline{\mathcal M}\right\}.
\end{equation}
The canonical projection $\pi:\overline{\mathcal M}\rightarrow\mathcal M$ is given by $\pi(x)=[x]$.
Equipped with a smooth structure, $\mathcal M$ is a quotient manifold of $\overline{\mathcal M}$ if $\pi$ is a smooth submersion, i.e., $\operatorname{rank}D\pi(x)=\dim\mathcal M$ for every $x\in\overline{\mathcal M}$, where $D\pi(x):T_x\overline{\mathcal M}\rightarrow T_{[x]}\mathcal M$ denotes the differential of $\pi$ at $x$.
The fiber through $x$ is $\mathcal F_x=\pi^{-1}\bigl(\pi(x)\bigr)$, which is an embedded submanifold of $\overline{\mathcal M}$.
Its tangent space at $x$ defines the vertical space $\mathcal V_x=T_x\mathcal F_x=\ker D\pi(x)$.
Consequently, two tangent vectors $\bm{u},\bm{v}\in T_x\overline{\mathcal M}$ represent the same tangent vector in $T_{[x]}\mathcal M$ if and only if their difference is vertical:
\begin{equation}
D\pi(x)[\bm{u}]=D\pi(x)[\bm{v}]\quad\Longleftrightarrow\quad \bm{u}-\bm{v}\in\mathcal V_x.
\end{equation}

\section{Method}

\begin{figure}[t]
\centering
\includegraphics[width=\textwidth]{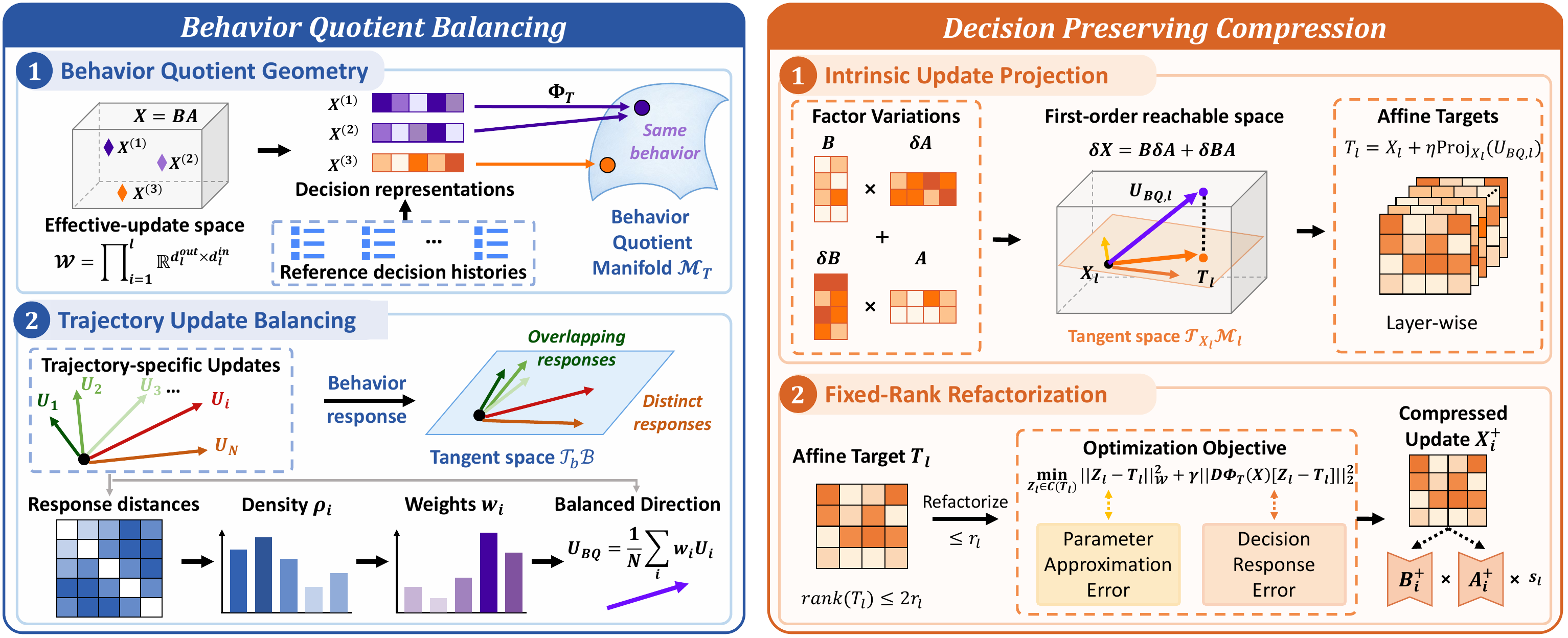}
\caption{Overview of \modelname. \moduleA~constructs a behavior quotient manifold and balances trajectory updates using quotient distances. \moduleB~projects the balanced direction onto the fixed rank LoRA tangent space and refactorizes the resulting target while preserving decision changes.}
\label{fig:Framework}
\end{figure}

\subsection{Problem Statement}

We consider an LLM-based agent with LoRA applied to $L$ weight matrices.
For each adapted weight matrix indexed by $l\in\{1,\ldots,L\}$, LoRA parameterizes the adapted weight as:
\begin{equation}
\bm W_l=\bm W_l^0+\bm X_l,\qquad \bm X_l=s_l\bm B_l\bm A_l.
\end{equation}
The matrices $\bm W_l^0$ and $\bm W_l$ are the frozen pretrained and adapted weights, respectively, and $\bm X_l$ is the effective LoRA update.
The update is parameterized by trainable factors $\bm A_l\in\mathbb R^{r_l\times d_l^{\mathrm{in}}}$ and $\bm B_l\in\mathbb R^{d_l^{\mathrm{out}}\times r_l}$, with scaling coefficient $s_l>0$.
Their inner dimension $r_l$ defines the fixed rank budget and ensures $\operatorname{rank}(\bm X_l)\leq r_l$.

An agent trajectory is denoted by $\tau_i=\left(x_i,\left(h_{ik},y_{ik},o_{ik}\right)_{k=1}^{T_i}\right)$, where $x_i$ is the task, $h_{ik}$ is the history before step $k$, $y_{ik}$ is the agent output, and $o_{ik}$ is the environment feedback.
Each output $y_{ik}$ consists of intermediate reasoning and a decision $a_{ik}$.
The decision either advances the interaction or serves as the final task output.
Given a collection of $N$ trajectories $\mathcal T=\{\tau_i\}_{i=1}^{N}$, we write $\bm X=(\bm X_l)_{l=1}^{L}$ for the effective updates and let $\ell_i(\bm X)$ denote the task loss associated with $\tau_i$.
Standard LoRA training minimizes the empirical task loss:
\begin{equation}
\mathcal L_{\mathrm{task}}(\bm X)=\frac{1}{N}\sum_{i=1}^{N}\ell_i(\bm X).
\end{equation}

\subsection{Framework Overview}
The overall framework of \modelname~is illustrated in Figure~\ref{fig:Framework}, showing how trajectory updates are balanced and incorporated into a single LoRA adapter.
It consists of two modules, i.e., behavior quotient balancing (\moduleA) and decision preserving compression (\moduleB).
First, \moduleA~constructs normalized decision representations at fixed interaction histories and identifies effective updates that produce the same representations, forming a local behavior quotient geometry.
Within this geometry, trajectory update directions are compared through their first-order effects on the reference decision representations.
Their local concentration determines their relative weights, reducing the dominance of repeatedly represented behavioral changes and increasing the contribution of less represented ones.
The weighted trajectory directions are then aggregated into a shared effective update direction.
Second, \moduleB~projects the balanced direction onto the fixed rank tangent space induced by the current LoRA factors and takes a step along it to form an effective-weight target.
It then compresses this target under the original rank budget by jointly minimizing weight approximation error and the distortion of first-order decision responses measured using the same behavior geometry.
The compressed target is refactorized into updated LoRA factors for the next training step.
These two stages are repeated during training to balance trajectory contributions while controlling the behavioral distortion introduced by the rank constraint.

\subsection{Behavior Quotient Balancing}

\paragraph{Behavior Quotient Geometry.}
Trajectory updates can differ substantially in parameter space while inducing the same change in agent decisions.
Their differences may include components that are invisible at the decision histories of interest, causing parameter-space distances to overstate behavioral diversity.
We therefore construct a behavior map from decision representations and identify effective updates that agree under this map.
Let $\mathcal W=\prod_{l=1}^{L}\mathbb R^{d_l^{\mathrm{out}}\times d_l^{\mathrm{in}}}$ be the effective-update space with the product Frobenius inner product, and let $f_{\bm X}(h)\in\mathbb R^{d_{\mathrm{hid}}}$ denote the final representation passed to the frozen output head at history $h$.
For each reference decision $a_{ik}$, let $|a_{ik}|$ count its generation positions and let $h_{ik,t}$ extend $h_{ik}$ with the portion of $y_{ik}$ preceding position $t$, for $1\leq t\leq |a_{ik}|$.
These prefixes include the recorded reasoning and preceding decision tokens, so each representation describes the model state from which a decision token is predicted.
We keep the prefixes fixed across effective updates and collect the resulting representations in the behavior map:
\begin{equation}
\left[\Phi_{\mathcal T}(\bm X)\right]_{ik,t}
=\frac{f_{\bm X}(h_{ik,t})}{\sqrt{NT_i|a_{ik}|}}.
\label{eq:behavior_map}
\end{equation}
The normalization averages squared representation responses over generation positions, decisions, and trajectories.
Using this map, we define behavioral equivalence on the reference collection:
\begin{equation}
\bm X^{(1)}\sim_{\mathcal T}\bm X^{(2)}
\quad\Longleftrightarrow\quad
\Phi_{\mathcal T}(\bm X^{(1)})=\Phi_{\mathcal T}(\bm X^{(2)}).
\label{eq:behavior_equivalence}
\end{equation}
On a sufficiently small open neighborhood $\mathcal O$ of the current iterate $\bm X$ where $\Phi_{\mathcal T}$ is smooth with constant differential rank, these equivalence classes form the behavior quotient manifold $\mathcal M_{\mathcal T}=\mathcal O/\sim_{\mathcal T}$.
The quotient map $\pi_{\mathcal T}(\bm X)=[\bm X]_{\mathcal T}$ assigns each effective update to its behavior class, and the Euclidean geometry of $\Phi_{\mathcal T}$ induces a Riemannian metric $g$ on $\mathcal M_{\mathcal T}$.
At the current iterate, the quotient tangent space identifies directions whose difference lies in $\ker D\Phi_{\mathcal T}(\bm X)$.
Such directions have the same first-order response on the reference decisions even when their effective-weight components differ.
This construction makes behavioral overlap measurable at the current model: invisible components contribute no quotient distance, while distinguishable decision responses remain separated.

\paragraph{Trajectory Update Balancing.}
The quotient geometry reveals overlap between trajectory updates, but uniform weighting still counts each trajectory as an independent contribution.
When many trajectories induce similar decision responses, their repeated contributions can dominate the aggregate direction.
Behavioral changes represented by fewer trajectories then receive less influence, even when they provide distinct responses on the reference collection.
We use the local concentration of quotient tangent directions to adjust these relative contributions.
For each trajectory, we take its task direction $\bm U_i=-\nabla_{\bm X}\ell_i(\bm X)$ and map it to the quotient tangent $D\pi_{\mathcal T}(\bm X)[\bm U_i]$ at $[\bm X]_{\mathcal T}$.
Its behavior response $D\Phi_{\mathcal T}(\bm X)[\bm U_i]$ describes how training on trajectory $i$ would change all reference decision representations to first order.
We compare two such directions under the quotient metric $g$:
\begin{equation}
d_{ij}^2
=\left\|D\pi_{\mathcal T}(\bm X)[\bm U_i-\bm U_j]\right\|_g^2
=\left\|D\Phi_{\mathcal T}(\bm X)[\bm U_i-\bm U_j]\right\|_2^2.
\label{eq:bqb_distance}
\end{equation}
The right-hand expression evaluates the quotient tangent distance directly from the two behavior responses.
It can be computed using Jacobian-vector products without explicitly constructing quotient coordinates.
We then use Gaussian affinities~\citep{DiffusionMaps} to accumulate the overlap surrounding each direction and set its relative weight:
\begin{equation}
\rho_i=\sum_{j=1}^{N}\exp\left(-\frac{d_{ij}^2}{2\sigma_{\mathcal T}^2}\right),
\qquad
w_i=\frac{\rho_i^{-\alpha}}{\frac{1}{N}\sum_{j=1}^{N}\rho_j^{-\alpha}}.
\label{eq:bqb_weights}
\end{equation}
Here $\sigma_{\mathcal T}>0$ is a data-dependent bandwidth that determines the scale of the neighborhood comparison.
A larger $\rho_i$ means that more of the trajectory collection produces responses close to that of $\bm U_i$.
The exponent $\alpha\in[0,1]$ controls how strongly this overlap reduces the relative weight of a trajectory.
The normalization keeps $N^{-1}\sum_iw_i=1$, preserving the total weight of the empirical objective, and $\alpha=0$ recovers uniform weighting.
For groups of similar responses with sufficiently small overlap between groups, $\alpha=1$ approximately equalizes their total contributions.
Holding the computed weights fixed during the task update, we combine the original trajectory directions:
\begin{equation}
\bm U_{\mathrm{BQ}}
=\frac{1}{N}\sum_{i=1}^{N}w_i\bm U_i.
\label{eq:bqb_update}
\end{equation}
By linearity of $D\pi_{\mathcal T}(\bm X)$, the aggregate induces the corresponding weighted mean of the trajectory quotient tangents.
For $\alpha>0$, this reduces the relative contribution of densely represented behavioral changes and increases that of less represented changes.
Each task loss continues to determine the sign and direction of its own update, while the positive weights control its contribution to the shared adapter.
BQB therefore adjusts behavioral coverage while retaining the original task supervision, without requiring predefined capability partitions.

\subsection{Decision Preserving Compression}

\paragraph{Intrinsic Update Projection.}
BQB aggregates trajectory directions in the effective-update space, whereas the stored LoRA adapter changes its weights through low-rank factors.
The balanced direction may consequently contain components that these factors cannot realize to first order at the current iterate.
To obtain a locally attainable update, we project it onto the space of effective directions induced by perturbing the factors.
This projection should depend on the current effective weights, so that choosing an equivalent factorization does not change the target.
For each layer $l$, let $\mathcal R_{r_l}$ denote the manifold of matrices in $\mathbb R^{d_l^{\mathrm{out}}\times d_l^{\mathrm{in}}}$ with rank $r_l$.
At an iterate with $\operatorname{rank}(\bm X_l)=r_l$, differentiating $\bm X_l=s_l\bm B_l\bm A_l$ yields the tangent space of effective directions attainable to first order through the LoRA factors~\citep{LoRA-Pro}:
\begin{equation}
T_{\bm X_l}\mathcal R_{r_l}=\left\{s_l\left(\bm B_l\Delta\bm A_l+\Delta\bm B_l\bm A_l\right):\Delta\bm A_l\in\mathbb R^{r_l\times d_l^{\mathrm{in}}},\ \Delta\bm B_l\in\mathbb R^{d_l^{\mathrm{out}}\times r_l}\right\}.
\end{equation}
The two terms describe the effective changes obtained by varying either factor, and their sum contains all first-order directions attainable at $\bm X_l$.
Let $\bm P_l\in\mathbb R^{d_l^{\mathrm{out}}\times r_l}$ and $\bm Q_l\in\mathbb R^{d_l^{\mathrm{in}}\times r_l}$ have orthonormal columns spanning $\operatorname{col}(\bm X_l)$ and $\operatorname{col}(\bm X_l^{\top})$, respectively.
We denote the Frobenius orthogonal projection onto $T_{\bm X_l}\mathcal R_{r_l}$ by $\operatorname{Proj}_{\bm X_l}$.
The projected layerwise direction is~\citep{FixedRankDNN}:
\begin{equation}
\operatorname{Proj}_{\bm X_l}\left(\bm U_{\mathrm{BQ},l}\right)=\bm P_l\bm P_l^{\top}\bm U_{\mathrm{BQ},l}+\bm U_{\mathrm{BQ},l}\bm Q_l\bm Q_l^{\top}-\bm P_l\bm P_l^{\top}\bm U_{\mathrm{BQ},l}\bm Q_l\bm Q_l^{\top}.
\end{equation}
The first two terms retain the components accessible through the current column or row space, while the last term subtracts their shared component.
With a learning rate $\eta>0$, the projected direction defines the layerwise affine target:
\begin{equation}
\bm T_l=\bm X_l+\eta\operatorname{Proj}_{\bm X_l}\left(\bm U_{\mathrm{BQ},l}\right).
\end{equation}
We collect these layerwise targets as $\bm T=(\bm T_l)_{l=1}^{L}$.
The Frobenius projection gives the closest locally attainable direction to $\bm U_{\mathrm{BQ},l}$, minimizing the alteration of the balanced direction within the tangent constraint.
Because the projectors $\bm P_l\bm P_l^{\top}$ and $\bm Q_l\bm Q_l^{\top}$ depend only on the column and row spaces of $\bm X_l$, the projected direction and the resulting target are invariant to equivalent LoRA factorizations.
This gives the compression stage a target defined by the effective weights of the adapter and its local ability to change them.
The first-order decision change induced by $\bm T-\bm X$ is the reachable target effect that the subsequent refactorization aims to preserve.

\paragraph{Fixed-Rank Refactorization.}
The tangent projection ensures that the update direction is attainable to first order, but a finite step along that direction need not remain on the fixed-rank manifold.
In particular, the affine target $\bm T_l$ can be expressed as the sum of two matrices of rank at most $r_l$, giving $\operatorname{rank}(\bm T_l)\leq 2r_l$.
When $\operatorname{rank}(\bm T_l)>r_l$, the target cannot be represented under the original LoRA rank budget and must be compressed.
Equal-sized weight errors can cause different changes in the reference decisions because the behavior map is more sensitive to some residual directions than others.
Weight-space approximation alone therefore does not distinguish which parts of the target matter most for its decision effect.
We use the behavior geometry from BQB to guide the selection of a rank-constrained representative.
Let $\mathcal C(\bm T)\subseteq\mathcal W$ contain candidate updates $\bm Z$ satisfying $\operatorname{rank}(\bm Z_l)\leq r_l$ for every layer.
Each candidate is further constrained by $\operatorname{col}(\bm Z_l)\subseteq\operatorname{col}(\bm T_l)$ and $\operatorname{col}(\bm Z_l^{\top})\subseteq\operatorname{col}(\bm T_l^{\top})$.
These constraints restrict the refactorization to the column and row spaces already present in the projected target.
Within this candidate set, we select the next effective update by jointly measuring weight error and behavioral distortion:
\begin{equation}
\bm X^{+}\in\operatorname*{arg\,min}_{\bm Z\in\mathcal C(\bm T)}\left\{\left\|\bm Z-\bm T\right\|_{\mathcal W}^{2}+\gamma\left\|D\Phi_{\mathcal T}(\bm X)[\bm Z-\bm T]\right\|_2^2\right\}.
\end{equation}
The first term measures the effective-weight error introduced by rank reduction.
For the second term, linearity gives
$D\Phi_{\mathcal T}(\bm X)[\bm Z-\bm T]
=D\Phi_{\mathcal T}(\bm X)[\bm Z-\bm X]-D\Phi_{\mathcal T}(\bm X)[\bm T-\bm X]$.
It thus compares the first-order decision responses induced by the candidate step and the target step at the same current iterate.
The reference histories and the differential remain fixed during refactorization, so this term measures the distortion caused by replacing $\bm T$ with $\bm Z$.
The coefficient $\gamma>0$ controls the tradeoff between effective-weight approximation and decision preservation.
Because the behavior response is evaluated across all adapted layers jointly, the criterion accounts for their combined effect on the reference decisions.
The restricted target spaces permit optimization through cores of size at most $2r_l\times2r_l$.
This objective favors representatives that approximate the target weights while retaining the decision changes identified by the same geometry used for trajectory balancing.
The rank constraints guarantee factors $\bm B_l^{+}$ and $\bm A_l^{+}$ satisfying $\bm X_l^{+}=s_l\bm B_l^{+}\bm A_l^{+}$.
The compressed target can be stored as a single LoRA adapter under the original rank budgets, with decision distortion included in the compression criterion.

\section{Experiments}

\subsection{Experimental Setups}

\paragraph{Datasets.}
We conduct experiments on AppWorld~\citep{appworld} and BrowseComp-Plus~\citep{BrowseCompPlus}.
\textbf{AppWorld} evaluates agents on multi-step tasks in a stateful world of simulated applications.
Given a natural language request, the agent invokes application APIs and responds to environment feedback to complete the task.
Programmatic tests verify task completion and the resulting application states.
\textbf{BrowseComp-Plus} evaluates deep search agents on difficult questions that require iterative retrieval and evidence synthesis.
Agents search a fixed document corpus containing human-verified supporting evidence.
For both benchmarks, we perform supervised fine-tuning on collected successful trajectories.

\begin{table}[t]
\centering
\caption{Task performance on AppWorld and BrowseComp-Plus (\%, $\uparrow$).}
\label{tab:main_results}
\small
\setlength{\tabcolsep}{2.5pt}
\renewcommand{\arraystretch}{1.08}
\begin{tabular*}{\textwidth}{@{\extracolsep{\fill}}cl*{8}{c}@{}}
\toprule
& & \multicolumn{4}{c}{AppWorld} & \multicolumn{4}{c}{BrowseComp-Plus} \\
\cmidrule(lr){3-6}\cmidrule(lr){7-10}
& & \multicolumn{2}{c}{Test-N} & \multicolumn{2}{c}{Test-C} & \multicolumn{2}{c}{BM25} & \multicolumn{2}{c}{Embedding} \\
\cmidrule(lr){3-4}\cmidrule(lr){5-6}\cmidrule(lr){7-8}\cmidrule(lr){9-10}
Model & Method & TGC & SGC & TGC & SGC & Accuracy & Recall & Accuracy & Recall \\
\midrule
\multirow{8}{*}{Qwen3.5-4B} & Base Agent & 27.98 & 10.71 & 11.99 & 2.16 & 20.00 & 22.43 & 36.67 & 44.77 \\
 & Full & 58.33 & 39.29 & 40.29 & 21.58 & 26.00 & 26.60 & 44.00 & 48.10 \\
\cmidrule(lr){2-10}
 & LoRA & 52.98 & 32.14 & 35.49 & 17.99 & 20.67 & 22.28 & 40.00 & 45.77 \\
 & LoRA-S & \underline{58.93} & 41.07 & 39.33 & 19.42 & 26.67 & 27.07 & 42.00 & 47.30 \\
 & MoRAgent & \textbf{60.12} & \underline{42.86} & 40.05 & 21.58 & 25.33 & 26.40 & \underline{43.33} & 48.30 \\
 & DART & 56.55 & 37.50 & 39.09 & 20.14 & \underline{28.00} & \underline{28.90} & \textbf{45.33} & \underline{49.10} \\
 & TopoCurate & 55.95 & 37.50 & \underline{41.01} & \underline{23.02} & 26.00 & 27.60 & \underline{43.33} & 48.80 \\
\addlinespace[2pt]
 & \modelname & \underline{58.93} & \textbf{44.64} & \textbf{43.65} & \textbf{24.46} & \textbf{28.67} & \textbf{29.30} & \textbf{45.33} & \textbf{50.10} \\
\midrule
\multirow{8}{*}{Qwen3.5-9B} & Base Agent & 36.31 & 19.64 & 23.02 & 10.79 & 24.00 & 24.08 & 44.67 & 44.20 \\
 & Full & 69.64 & 53.57 & 49.64 & 27.34 & 30.00 & 28.40 & 49.33 & 48.20 \\
\cmidrule(lr){2-10}
 & LoRA & 60.71 & 42.86 & 45.08 & 20.86 & 26.00 & 24.83 & 46.00 & 45.20 \\
 & LoRA-S & 66.67 & \underline{51.79} & 47.72 & 25.18 & 29.33 & 27.60 & 48.00 & 47.50 \\
 & MoRAgent & \textbf{69.05} & \textbf{55.36} & 49.16 & 28.06 & 29.33 & 28.10 & 49.33 & 49.10 \\
 & DART & 66.67 & 48.21 & 48.20 & 25.90 & \underline{31.33} & \underline{30.20} & \underline{50.67} & \underline{49.70} \\
 & TopoCurate & 64.88 & 48.21 & \underline{50.12} & \underline{29.50} & 29.33 & 29.10 & 49.33 & 49.30 \\
\addlinespace[2pt]
 & \modelname & \underline{68.45} & \textbf{55.36} & \textbf{53.00} & \textbf{31.65} & \textbf{32.67} & \textbf{31.10} & \textbf{51.33} & \textbf{50.90} \\
\bottomrule
\end{tabular*}
\end{table}

\paragraph{Baselines.}
We compare \modelname~with seven baselines.
Standard adaptation baselines include:
(1) Base, which evaluates the pretrained model without adaptation,
(2) Full FT, which updates all model parameters,
and (3) LoRA~\citep{LoRA}, which introduces low-rank updates into each adapted layer.
We then compare two multi-LoRA methods designed for agent tuning:
(4) MoRAgent~\citep{MoRAgent}, which assigns separate LoRA groups to the reasoner, executor, and summarizer roles,
and (5) DART~\citep{DART}, which separates reasoning and tool use updates across distinct low-rank modules.
For single-LoRA optimization, we include
(6) LoRA-S~\citep{LoRA-S}, which optimizes low-rank factors on a quotient manifold through a Sylvester equation.
For trajectory curation, we include
(7) TopoCurate~\citep{TopoCurate}, which uses a semantic quotient topology of multi-trial interactions to select informative trajectories.


\subsection{Performance Evaluation}

\paragraph{Main Results.}
Tables~\ref{tab:main_results} and~\ref{tab:interaction_counts} summarize task performance and interaction counts, respectively.
We make the following observations:
(1) Standard LoRA improves on the base agent, but its gains are limited on evidence retrieval and it remains well below the stronger baselines.
This gap suggests that standard low-rank adaptation leaves room for better allocation of adapter capacity across agent behaviors.
(2) LoRA-S consistently outperforms standard LoRA, supporting the value of eliminating redundant degrees of freedom in the low-rank factorization.
However, its equivalence relation identifies factorizations of the same effective weights and does not address redundancy between trajectory-induced behavioral changes.
(3) Multiple adapters provide additional benefits through capability specialization.
MoRAgent performs best on Test-N among the low-rank baselines, while DART leads these baselines on BrowseComp-Plus.
However, these gains rely on separate adapters and predefined role or capability partitions, and their relative strengths vary across tasks.
(4) TopoCurate improves trajectory selection by reducing semantic repetition and retaining informative interactions, yielding the strongest baseline performance on Test-C.
However, similarity between recorded actions and observations does not determine whether optimizing two trajectories will induce the same behavioral change.
Its curation criterion therefore leaves redundancy in the resulting behavioral changes unresolved.
(5) Averaged over the eight reported performance metrics, \modelname~outperforms the strongest baseline, MoRAgent, by $2.14$ and $2.12$ percentage points for the 4B and 9B backbones, respectively.
It also achieves the lowest interaction counts in all evaluated settings.
Balancing trajectory updates and preserving their decision effects during compression enable strong performance and efficient interaction in a single fixed-rank adapter.

\begin{table}
\centering
\caption{Mean interaction counts on AppWorld (steps) and BrowseComp-Plus (search calls) ($\downarrow$).}
\label{tab:interaction_counts}
\small
\setlength{\tabcolsep}{4pt}
\renewcommand{\arraystretch}{1.08}
\begin{tabular*}{\textwidth}{@{\extracolsep{\fill}}cl*{4}{c}@{}}
\toprule
& & \multicolumn{2}{c}{AppWorld} & \multicolumn{2}{c}{BrowseComp-Plus} \\
\cmidrule(lr){3-4}\cmidrule(lr){5-6}
& & Test-N & Test-C & BM25 & Embedding \\
Model & Method & Avg. Steps & Avg. Steps & Search Calls & Search Calls \\
\midrule
\multirow{8}{*}{Qwen3.5-4B} & Base Agent & 31.28 & 36.64 & 16.95 & 15.54 \\
 & Full & 17.75 & 26.10 & 16.70 & 11.65 \\
\cmidrule(lr){2-6}
 & LoRA & 18.26 & 27.12 & 17.73 & 14.35 \\
 & LoRA-S & 18.65 & 27.20 & 17.20 & 13.25 \\
 & MoRAgent & \underline{17.10} & 25.35 & 16.85 & 12.35 \\
 & DART & 18.00 & 26.25 & 16.15 & \underline{11.90} \\
 & TopoCurate & 17.35 & \underline{24.90} & \underline{15.75} & 12.05 \\
\addlinespace[2pt]
 & \modelname & \textbf{16.70} & \textbf{24.10} & \textbf{15.35} & \textbf{11.35} \\
\midrule
\multirow{8}{*}{Qwen3.5-9B} & Base Agent & 26.20 & 29.92 & 15.65 & 13.16 \\
 & Full & 17.45 & 24.25 & 15.50 & 11.10 \\
\cmidrule(lr){2-6}
 & LoRA & 17.16 & 24.96 & 17.92 & 13.48 \\
 & LoRA-S & 17.65 & 25.75 & 16.95 & 12.65 \\
 & MoRAgent & \underline{16.10} & 23.60 & 16.30 & 11.80 \\
 & DART & 16.90 & 24.20 & 15.75 & \underline{11.55} \\
 & TopoCurate & 16.65 & \underline{23.15} & \underline{15.30} & 11.65 \\
\addlinespace[2pt]
 & \modelname & \textbf{15.95} & \textbf{22.70} & \textbf{15.05} & \textbf{10.85} \\
\bottomrule
\end{tabular*}
\end{table}

\paragraph{Ablation Study.}
Table~\ref{tab:ablation} evaluates variants with each module removed.
(1) Removing \moduleA~reduces Test-C performance by $3.57$ and $3.83$ percentage points for the 4B and 9B backbones, respectively, and lowers BrowseComp-Plus accuracy with both retrievers.
This supports balancing trajectory updates by their effects on decision representations.
The reduction is larger on Test-C than on Test-N for both backbones, indicating that the benefit of balancing varies across evaluation settings.
(2) Removing \moduleB~reduces performance in all eight settings, with average drops across the four metrics of $3.64$ and $3.57$ percentage points for the 4B and 9B backbones, respectively.
This supports preserving decision changes when compressing the balanced update under the fixed rank budget.
The consistent drops across AppWorld and both BrowseComp-Plus retrievers suggest that decision preservation contributes across different forms of agent interaction.
(3) The full method leads in seven of eight settings. Without \moduleA, the 4B model scores $0.20$ percentage points higher on Test-N.
The two modules provide complementary benefits, although balancing does not improve every setting.
Both partial variants also outperform standard LoRA in every setting, indicating that each component retains useful gains when applied without the other.
The ranking of the partial variants is consistent across backbones: retaining \moduleB~gives higher mean performance than retaining \moduleA~alone.
Relative to standard LoRA, the full method improves the mean of the four ablation metrics by $6.86$ and $6.92$ percentage points for the 4B and 9B backbones, respectively.

\begin{table}
\centering
\caption{Ablation results on AppWorld (TGC) and BrowseComp-Plus (accuracy) (\%, $\uparrow$).}
\label{tab:ablation}
\small
\setlength{\tabcolsep}{2pt}
\renewcommand{\arraystretch}{1.12}
\begin{tabular*}{\textwidth}{@{\extracolsep{\fill}}cl*{4}{c}@{}}
\toprule
& & \multicolumn{2}{c}{AppWorld} & \multicolumn{2}{c}{BrowseComp-Plus} \\
\cmidrule(lr){3-4}\cmidrule(lr){5-6}
Model & Method & Test-N & Test-C & BM25 & Embedding \\
\midrule
\multirow{4}{*}{Qwen3.5-4B} & LoRA & $52.98\pm1.10$ & $35.49\pm1.45$ & $20.67\pm1.30$ & $40.00\pm1.00$ \\
\cmidrule(lr){2-6}
 & \modelname~w/o BQB & $\mathbf{59.13\pm1.00}$ & $\underline{40.08\pm1.60}$ & $\underline{27.07\pm1.40}$ & $\underline{43.10\pm1.15}$ \\
 & \modelname~w/o DPC & $55.38\pm1.35$ & $39.59\pm1.75$ & $24.67\pm1.60$ & $42.40\pm1.35$ \\
\cmidrule(lr){2-6}
 & \modelname & $\underline{58.93\pm1.05}$ & $\mathbf{43.65\pm1.40}$ & $\mathbf{28.67\pm1.25}$ & $\mathbf{45.33\pm1.05}$ \\
\midrule
\multirow{4}{*}{Qwen3.5-9B} & LoRA & $60.71\pm1.00$ & $45.08\pm1.30$ & $26.00\pm1.20$ & $46.00\pm0.95$ \\
\cmidrule(lr){2-6}
 & \modelname~w/o BQB & $\underline{67.67\pm0.90}$ & $49.17\pm1.35$ & $\underline{30.13\pm1.25}$ & $\underline{49.10\pm1.00}$ \\
 & \modelname~w/o DPC & $63.51\pm1.20$ & $\underline{49.78\pm1.55}$ & $29.20\pm1.45$ & $48.70\pm1.15$ \\
\cmidrule(lr){2-6}
 & \modelname & $\mathbf{68.45\pm1.00}$ & $\mathbf{53.00\pm1.25}$ & $\mathbf{32.67\pm1.15}$ & $\mathbf{51.33\pm0.95}$ \\
\bottomrule
\end{tabular*}
\end{table}

\noindent\begin{minipage}{\linewidth}
\paragraph{Hyperparameter Sensitivity.}
We study the balancing exponent $\alpha$ and compression coefficient $\gamma$.
We test $\alpha\in\{0,0.25,0.5,0.75,1\}$ and $\gamma\in\{0.1,0.3,0.5,0.8,1,3,10\}$, varying one coefficient at a time while fixing the other at $1$.
Figure~\ref{fig:hyperparameter_sensitivity} reports averages over the eight performance metrics for
\begin{wrapfigure}{r}{0.5\textwidth}
\centering
\includegraphics[width=\linewidth]{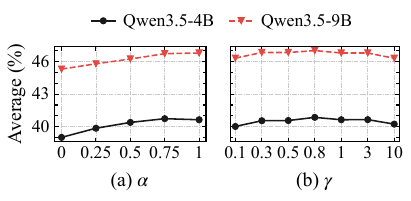}
\setlength{\abovecaptionskip}{4pt}
\caption{Hyperparameter sensitivity.}
\label{fig:hyperparameter_sensitivity}
\end{wrapfigure}
both backbones.
(1) Performance peaks at ${\alpha=0.75}$ for 4B and $\alpha=1$ for 9B.
At $\alpha=0$, all trajectories receive equal weight, providing a reference without density-based balancing.
Increasing $\alpha$ reduces the relative contribution of densely represented responses.
Both backbones improve over uniform weighting, with the 4B curve flattening near the upper end and the 9B curve continuing to rise.
(2) Both backbones perform best at $\gamma=0.8$, with nearby values remaining competitive.
A small $\gamma$ gives greater relative emphasis to weight approximation, while a large $\gamma$ increases the penalty on local decision distortion.
The effective-weight error remains part of the objective at every tested value, so this sweep varies the balance between the two compression criteria.
Performance decreases toward both ends of the tested range, supporting an intermediate tradeoff between weight fidelity and decision preservation.
This pattern holds across a sweep spanning two orders of magnitude in $\gamma$ and is shared by both backbone sizes.
\end{minipage}

\section{Conclusion}
We presented \modelname~for adapting LLM agents with a single fixed-rank LoRA.
\moduleA~balances trajectory updates in a local behavior quotient geometry, reducing repeated contributions from updates with equivalent effects on decision representations.
\moduleB~projects the balanced update and compresses the resulting target while controlling decision distortion.
Together, they preserve the original rank budget and require no capability-specific routing.
Experiments on AppWorld and BrowseComp-Plus show strong task performance and reduced interaction counts.

\clearpage
\bibliography{iclr2027_conference}
\bibliographystyle{iclr2027_conference}

\end{document}